\documentclass[runningheads]{llncs}
\usepackage[T1]{fontenc}
\usepackage{graphicx,verbatim}
\usepackage{mwe} 
\usepackage{amssymb}
\usepackage{pifont} 
\usepackage{booktabs}
\usepackage{hyperref}
\usepackage{graphicx}
\usepackage{xcolor}
\usepackage[table]{xcolor}
\usepackage{makecell}
\usepackage{caption}   
\usepackage{booktabs}
\usepackage[table]{xcolor}
\usepackage{amssymb}
\usepackage{amsmath}
\usepackage{xurl}        
\definecolor{olivegreen}{RGB}{20,140,47}
\definecolor{custombrown}{RGB}{200,30,0}

\begin{document}
 \title{How Good are Foundation Models in Longitudinal MRI Disease Progression Reasoning?}
 
 \titlerunning{Time-Aware Multi-View MRI Benchmark}

%

\author{
Wafa Al Ghallabi\inst{1} \and
Ritesh Thawkar\inst{1} \and
Sara Ghaboura\inst{1} \and
Omkar Thawakar\inst{1} \and
Numan Saeed\inst{1} \and
Dana Al Nuaimi\inst{2} \and
Ajnas Alkatheeri\inst{3} \and
Salman Khan\inst{1} \and
Fahad Shahbaz Khan\inst{1,4}
}
%
%
\authorrunning{W. Al Ghallabi et al.}
\institute{
Mohamed bin Zayed University of Artificial Intelligence, Abu Dhabi, UAE \\
\email{wafa.alghallabi@mbzuai.ac.ae} \and
Department of Health Abu Dhabi, Abu Dhabi, UAE \and
Fatima College of Health Sciences, Abu Dhabi, UAE \and
Link\"oping University, Link\"oping, Sweden
}

  
\maketitle              
\begin{abstract}
Magnetic Resonance Imaging (MRI) interpretation is fundamental to clinical decision-making, requiring radiologists to integrate multi-view anatomical planes across sequential timepoints while precisely localizing interval changes. However, existing vision-language benchmarks remain confined to single-timepoint, single-view interpretation, failing to capture the temporal-spatial reasoning essential to radiologic practice. We introduce the \textit{Time-Aware Multi-View MRI Benchmark}, an evaluation framework unifying multi-view anatomical input, temporal reasoning across longitudinal scans, and structured localization guidance. The benchmark comprises 3{,}920 expert-verified question-answer pairs derived from 890 patients across over 3{,}200 longitudinal MRI timepoints, drawn from seven clinical cohorts covering glioblastoma, neurodegeneration, vestibular schwannoma, and brain metastases, in open-ended, multiple-choice, and binary formats, requiring models to identify anatomical regions of maximal change, characterize progression across sequences and views, and provide structured guidance specifying boundaries, imaging features, and confounders. Experiments across 16 vision-language models reveal moderate temporal alignment but systematic failure on change direction recognition and volumetric quantification, while multi-view inputs improve spatial localization yet degrade temporal reasoning in compact architectures. Our benchmark provides a systematic framework for evaluating progression tracking, interval change localization, and temporal ordering, which are essential for clinical deployment.
Code, evaluation splits, and the dataset are available at~\url{https://github.com/wafaAlghallabi/Time-Aware-MRI}.
\end{abstract}

\section{Introduction}

Magnetic Resonance Imaging (MRI) serves as the cornerstone of longitudinal disease assessment, enabling clinicians to evaluate treatment response, detect recurrence, and characterize the temporal trajectory of pathology. Clinical interpretation of follow-up MRI is inherently comparative and multi-dimensional. Radiologists systematically align current and prior studies across multiple anatomical planes (axial, coronal, sagittal), assess interval changes in lesion morphology and signal characteristics, and synthesize these observations into conclusions about disease progression~\cite{acr_practice_parameter_2024}. This integrative reasoning, which combines spatial understanding across anatomical views, temporal comparison between timepoints, and precise localization of change, represents the cognitive foundation of radiologic decision-making. Despite this clinical reality, existing vision-language benchmarks evaluate models almost exclusively on static, single-timepoint interpretation.

\begin{figure}[t!]
\centering
\caption{Comparison of reasoning dimension coverage across MRI benchmarks. Time-Aware Multi-View MRI benchmark achieves full coverage across all five tasks. MedAtlas~\cite{medatlas2025} and OmniMRI~\cite{omnimri2025} offer partial coverage (p$^*$) on Temporal Reasoning and Disease Progression, respectively. Remaining benchmarks, NOVA~\cite{nova2025}, DrVD-Bench~\cite{drvd2025}, OmniMedVQA~\cite{omnimedvqa2024}, RadVUQA~\cite{radvuqa2025}, and MedFrameQA~\cite{medframeqa2025} provide no coverage across any dimension.}
\includegraphics[width=0.80\textwidth]{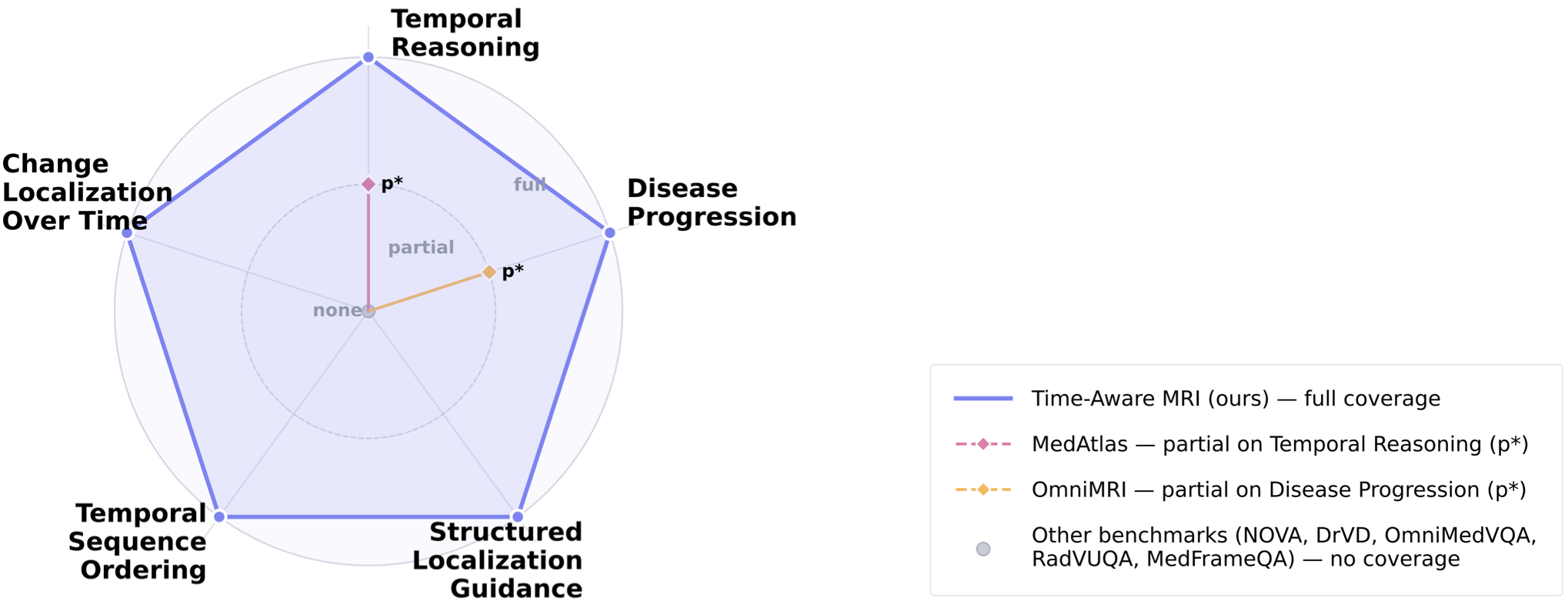}
\label{fig:benchmark_overview}
\end{figure}

Recent multimodal benchmarks have advanced medical image understanding but remain fundamentally limited to single-image analysis. Large-scale efforts such as OmniMedVQA~\cite{omnimedvqa2024}, GMAI-MMBench~\cite{gmai_mmbench_2024}, and PMC-VQA~\cite{pmc_vqa_2023} aggregate extensive collections but process each image independently. Limited efforts have begun addressing temporal reasoning~\cite{medical_diff_vqa_2025,medmim2025}, but explicitly acknowledge that longitudinal tracking and multi-view anatomical understanding remain unexplored. Consequently, despite impressive accuracy on static medical question-answering~\cite{med_gemini_2024}, vision-language models (VLMs) perform poorly on temporal tasks~\cite{temmed_bench_2025}, exposing fundamental training gaps. Furthermore, existing benchmarks lack structured localization guidance, including anatomical boundaries, imaging features, and confounders, which is essential for connecting semantic reasoning to spatial precision.

The disconnect between benchmark design and clinical requirements is compounded by documented failures of VLMs in spatial-temporal reasoning. Recent work highlights systematic failures in spatial localization and relative positioning~\cite{your_other_left_miccai2025}, with reports often lacking explicit anatomical references~\cite{radfm_natcomm_2025}. VLMs processing isolated slices without multi-view context may exhibit substantial performance degradation. Existing work incorporating temporal information~\cite{biovilt_cvpr2023} lacks multi-view anatomical analysis and segmentation capabilities, while work achieving multi-view reasoning~\cite{mosaic_cviu2025} demonstrates no temporal capability. Recent approaches~\cite{maira2_microsoft2024} combining viewing angles with temporal context operate exclusively on 2D radiography, unable to process axial-coronal-sagittal planes essential to cross-sectional imaging.

To address these gaps, we introduce temporal localization reasoning tasks that require models to identify anatomical regions of maximal interval change across viewing planes and provide structured guidance specifying boundaries, imaging features, and confounders to exclude. These tasks extend visual question answering beyond categorical recognition to integrated spatial-temporal localization, testing the radiological workflow from change detection to actionable delineation essential for treatment planning and longitudinal monitoring. The benchmark targets five tasks: Temporal Reasoning, Disease Progression, Structured Localization Guidance, Temporal Sequence Ordering, and Change Localization Over Time. 
\textbf{Our contributions are three-fold:}
\begin{itemize}
   \item \textbf{Novel Benchmark:} We present the Time-Aware Multi-View MRI Benchmark, a multi-disease evaluation framework for temporal-spatial reasoning in longitudinal neuroimaging. It integrates seven public cohorts (Glioblastoma, Alzheimer's, Vestibular Schwannoma, Brain Metastases) harmonized under a unified temporal protocol.
    \item \textbf{Structured Clinical Tasks:} We introduce tasks providing temporally aligned multi-sequence MRI across multi-view planes (axial, coronal, sagittal). Unlike prior categorical question answering (QA), our tasks require models to perform interval change localization with structured guidance (identifying boundaries, features, and confounders).
    \item \textbf{Comprehensive Evaluation:} We explicitly measure capabilities including progression tracking and temporal sequence ordering, providing systematic evaluation of five integrated dimensions absent from existing benchmarks (see Fig.~\ref{fig:benchmark_overview}).
\end{itemize}

\section{Time-Aware Multi-View MRI Benchmark}

\noindent\subsection{Dataset Composition and Preprocessing}

Our benchmark integrates seven longitudinal MRI cohorts selected for temporal depth and pathological diversity. The \textbf{Yale Brain Metastases Dataset}~\cite{yale_brainmets_2025} provides serial post-treatment monitoring documenting response to surgery and radiotherapy. Three glioblastoma cohorts capture distinct disease phases: the \textbf{UCSF Post-Operative Glioma Dataset}~\cite{rudie2024ucsf} tracks early post-surgical evolution, the \textbf{UCSD-PTGBM Dataset}~\cite{gagnon2026ucsd} includes rich molecular characterization (MGMT, IDH status) enabling correlation with imaging findings, and the \textbf{LUMIERE Dataset}~\cite{suter2022lumiere} provides expert RANO-annotated longitudinal scans across treatment cycles. The \textbf{OASIS-2}~\cite{oasis2_2010} and \textbf{ADNI}~\cite{adni2008} cohorts enable assessment of neurodegenerative progression in aging and Alzheimer's disease. The \textbf{Vestibular Schwannoma Multi-Center Dataset}~\cite{vsmcrc2023} contributes benign tumor monitoring data with multi-year follow-up intervals. Across cohorts, available sequences include T1, T2, FLAIR, T1CE, DWI, and ADC, with 3--4 timepoints per patient and inter-scan intervals ranging from 4 months in post-treatment glioma cohorts to over 18 months in vestibular schwannoma follow-up.

\begin{figure}[t!]
    \centering
    \caption{Time-Aware Multi-View MRI Benchmark pipeline: data selection, multi-view extraction, expert-guided question generation, and verification workflow.}
    \includegraphics[width=\textwidth]{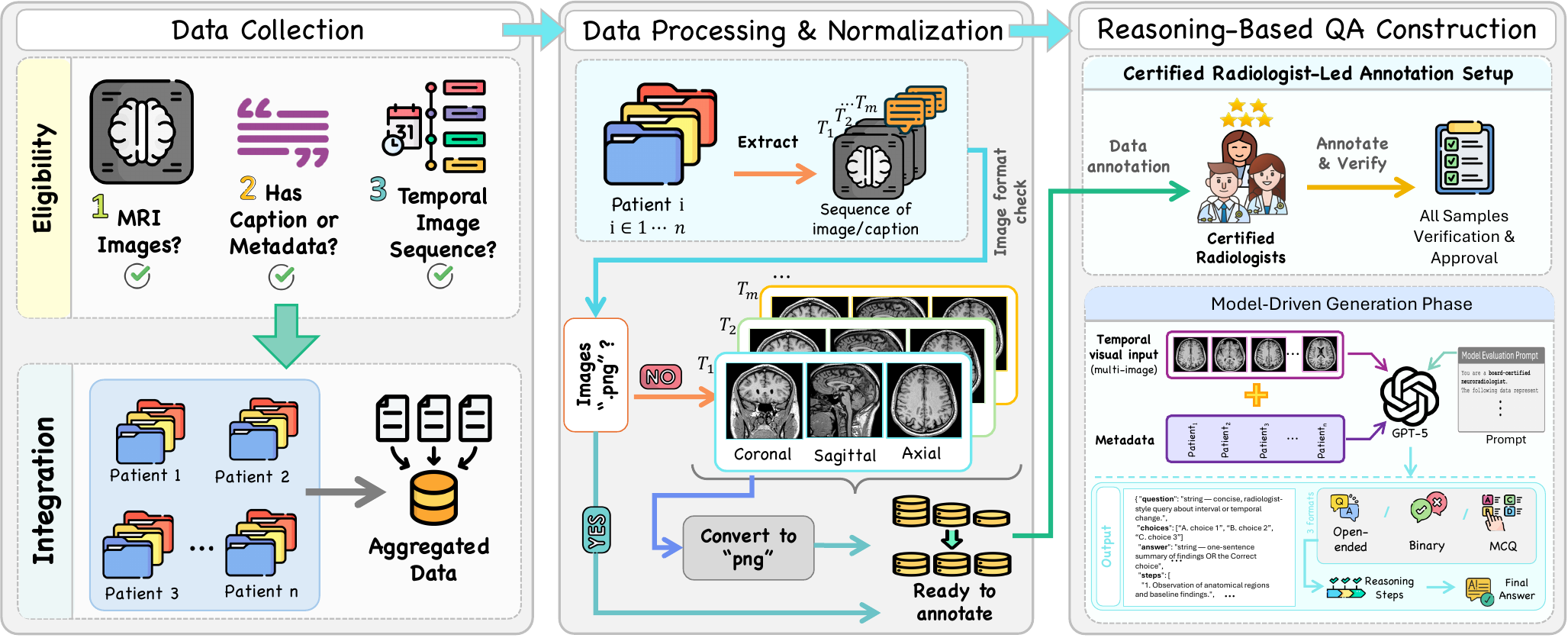}
    \label{fig:main}
\end{figure}

\noindent\textbf{Data Preprocessing and Quality Control.} To ensure cross-cohort consistency and clinical-grade image quality, we implemented a multi-stage preprocessing pipeline. Serial scans were registered to baseline using ANTs~\cite{avants2011ants} (deformable for most cohorts, rigid for geometrically consistent acquisitions). All volumes were reoriented to RAS+ neurological convention to ensure standardized anatomical presentation. For tumor cohorts, we employed tumor-aware slice extraction; segmentation masks identified tumor center of mass, and representative slices were extracted through the tumor centroid to guarantee lesion visibility in all views.
N4 bias field correction~\cite{tustison2010n4itk} was applied where needed, followed by sequence-specific percentile-based intensity normalization. To preserve clinically meaningful hyper- and hypointense signal characteristics critical for pathology assessment (e.g., enhancing tumor margins on T1CE, FLAIR hyperintensities), normalization windows were calibrated per sequence: T1CE and post-contrast sequences used $p_1$ to $p_{99.5}$ to retain enhancement extremes, while T2 and FLAIR used $p_2$ to $p_{98}$ with adaptive ceiling extension when bright lesion signal exceeded the upper bound. An automated quality control pipeline verified sufficient contrast, tumor visibility, and adequate tissue coverage. All registrations underwent dual expert review by two certified radiologists, with cases exhibiting motion artifacts, registration failures, or quality violations excluded. Temporal metadata including inter-scan intervals ($\Delta t$) and sequence identifiers were encoded to support time-aware reasoning.

\subsection{Multi-View and Temporal Integration}
Mainstream VLMs are constrained to 2D inputs~\cite{neurips2024_vote_mi}, limiting their application to volumetric imaging despite emerging 3D architectures~\cite{med3dvlm2025,merlin2024}. Radiologists, however, interpret MRI across orthogonal planes to triangulate findings~\cite{acr_practice_parameter_2024}. To approximate this workflow while remaining compatible with deployed VLMs, we extract multi-view presentations (axial, coronal, sagittal) from registered volumes. Representative slices across all available sequences (T1, T2, FLAIR, T1CE, DWI, ADC) yield 9-12 images per timepoint, and 3--4 serial timepoints per patient (months to years) enable longitudinal rather than single-scan evaluation.

\subsection{Question Generation and Expert Validation}
We developed a hybrid annotation pipeline combining automated generation with radiologist oversight (Fig.~\ref{fig:main}). For each longitudinal patient case, GPT-5 received the temporal MRI inputs and a structured prompt encoding cohort identity, temporal metadata ($\Delta t$, sequence parameters), available sequences, and clinical annotations where available (RANO scores, MGMT/IDH status, tumor volumes). This metadata grounding ensures questions reference verifiable findings rather than free-form image descriptions. Candidates were produced in three formats: open-ended for Temporal Reasoning and Disease Progression, multiple-choice with clinically plausible distractors for Structured Localization Guidance and Change Localization Over Time, and binary for Temporal Sequence Ordering. Each candidate included the question, options, final answer, and intermediate reasoning steps.

Two board-certified radiologists independently verified diagnostic accuracy, temporal consistency, distractor plausibility, and clinical relevance. Dual-approved samples were retained; flagged or conflicting items were returned to the generation pipeline with correction notes (e.g., ambiguous temporal reference, implausible distractor) and re-reviewed, with persistent disagreements discarded. This process yielded a 72\% acceptance rate and 3,920 expert-verified QA pairs.

\begin{figure}[t!]
    \centering
    \caption{Representative samples from the benchmark illustrating diverse pathologies and disease progression across serial timepoints (axial view).}
    \includegraphics[width=1\textwidth, height=3.25cm]{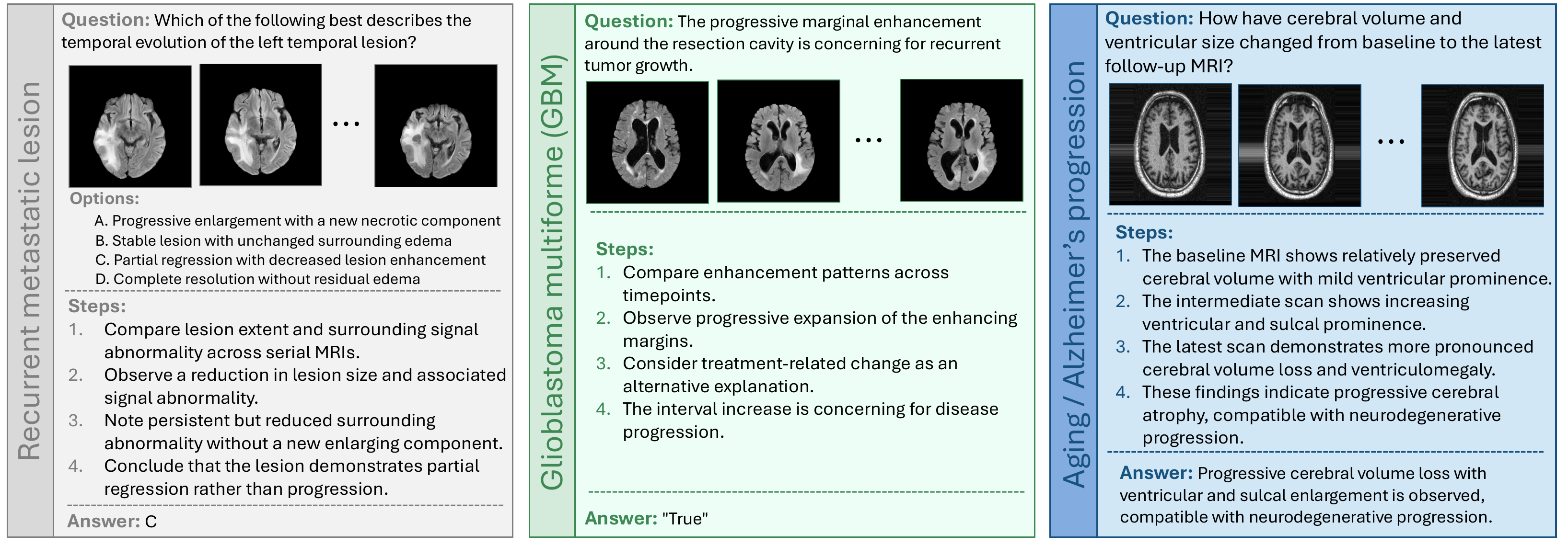}
    \label{fig:samples}
\end{figure}

\noindent\textbf{Benchmark Statistics.} The final benchmark comprises 3,920 QA pairs from 890 patients across over 3,200 timepoints (7.4 months mean follow-up), with disease distribution spanning glioblastoma (41\%), brain metastases (31\%), neurodegeneration (14\%), and vestibular schwannoma (14\%). Questions span five categories: (1)~\textit{Temporal Reasoning} (open-ended, 1,101 pairs): interval change identification, (2)~\textit{Disease Progression} (open-ended, 942): trajectory and treatment response, (3)~\textit{Structured Localization Guidance} (multiple-choice, 828): anatomical change regions with boundaries and imaging features, (4)~\textit{Temporal Sequence Ordering} (binary, 487): chronological reconstruction, and (5)~\textit{Change Localization Over Time} (multiple-choice, 562): maximal-change timepoints and locations. For localization tasks, structured guidance specifies boundaries, imaging features, and confounders to test full radiological workflow. Representative samples are shown in Fig.~\ref{fig:samples}.

\section{Evaluation Framework}

\noindent \textbf{Evaluation Tasks:} The \textit{Time-Aware Multi-View MRI Benchmark} evaluates model's ability to reason about disease evolution across serial multi-sequence, multi-view MRI through five complementary tasks. \textit{Temporal Reasoning} requires models to interpret sequential scans across multiple anatomical planes and imaging contrasts to answer clinically grounded questions about interval changes. \textit{Disease Progression} assesses the characterization of the overall disease trajectory (growth, regression, stability). \textit{Structured Localization Guidance} tests whether models can identify anatomical regions exhibiting change and provide structured descriptions of boundaries, imaging features, and relevant confounders. \textit{Temporal Sequence Ordering} evaluates the ability to correctly arrange scans chronologically based on visible progression patterns. Finally, \textit{Change Localization Over Time} requires precise identification of timepoints and spatial regions showing maximal interval change from baseline. Together, these tasks capture complementary dimensions of longitudinal understanding from spatial-temporal alignment and multi-sequence integration to biological interpretation, forming a unified framework for assessing temporal reasoning in longitudinal MRI.

\noindent \textbf{Evaluation Metrics:} To ensure fair assessment across diverse temporal reasoning tasks, we employ three complementary measures: \textit{Final Answer Accuracy}, \textit{Reasoning Score (RS)}, and the \textit{Time-Aware Composite (TAC)} metric. Final Answer Accuracy captures end-to-end correctness of model predictions. RS, computed following LlamaV-o1~\cite{thawakar2025llamavo1}, jointly considers reasoning consistency, temporal alignment, and factual accuracy to reflect how coherently models interpret longitudinal progression.

The TAC metric provides a unified measure of stepwise temporal reasoning, defined as:
\begin{equation}
\begin{split}
\label{eq:tac}
\mathrm{TAC} &= 0.5 \times \mathrm{TEDS} + 0.2 \times \mathrm{Trend\text{-}F1} \\
             &\quad + 0.2 \times \mathrm{SignAcc} + 0.1 \times \mathrm{Coverage}
\end{split}
\end{equation}

TAC integrates four components: \textit{Temporal Event Dependency Score (TEDS)} assesses alignment between predicted and reference change sequences while penalizing skipped or hallucinated intervals; \textit{Trend-F1} quantifies accuracy in detecting progression versus regression; \textit{Sign Accuracy} captures correctness of predicted change direction; and \textit{Coverage} reflects completeness of interval reasoning. Additionally, a \textit{Chronology} metric evaluates whether model-generated narratives preserve the correct temporal ordering of follow-up events. Together, these metrics assess not only whether models detect change, but how coherently they reason about timing, direction, and sequence across multi-view, multi-sequence temporal data. TAC ranges from 0 to 1, with higher scores indicating stronger temporal consistency and reasoning fidelity.

\begin{table*}[t!]
\centering
\caption{Model performance on the \textit{Time-Aware Multi-View MRI Benchmark} across final accuracy, Reasoning Score (RS), and Time-Aware Composite (TAC) metric. Higher TAC indicates stronger temporal consistency and reasoning fidelity. Bold: best per column.}
\label{tab:results}
\resizebox{\linewidth}{!}{
\begin{tabular}{lcccccccc}
\toprule
\textbf{Model} & \textbf{Final Acc (\%)} & \textbf{RS} & \textbf{TAC} & \textbf{TEDS} & \textbf{Trend F1} & \textbf{Sign Acc} & \textbf{Coverage} & \textbf{Chronology} \\
\midrule
\multicolumn{9}{c}{\textit{Closed-source models}} \\
\midrule
o4-mini & 32.18 & \textbf{6.68} & 0.753 & 0.832 & 0.548 & 0.681 & 0.908 & 0.918 \\
GPT-4o & 32.00 & 6.26 & 0.731 & 0.807 & 0.546 & 0.654 & 0.877 & 0.917 \\
GPT-5.2 & 21.20 & 5.83 & 0.661 & 0.805 & 0.192 & 0.639 & 0.921 & 0.856 \\
Gemini-2.5-Flash & 23.57 & 5.83 & 0.692 & 0.780 & 0.477 & 0.596 & 0.875 & 0.825 \\
Gemini-2.5-Pro & 23.66 & 5.88 & 0.672 & 0.785 & 0.504 & 0.528 & 0.730 & 0.957 \\
Gemini-3-Flash & 22.30 & 5.17 & 0.577 & 0.764 & 0.216 & 0.470 & 0.575 & \textbf{1.000} \\
Gemini-3-Pro & 35.10 & 5.31 & 0.590 & 0.775 & 0.235 & 0.485 & 0.600 & 0.980 \\
\midrule
\multicolumn{9}{c}{\textit{Open-source models}} \\
\midrule
InternVL3.5-Inst & \textbf{35.15} & 6.68 & \textbf{0.800} & \textbf{0.870} & \textbf{0.631} & \textbf{0.740} & 0.903 & 0.951 \\
Qwen3-VL-Plus-Thinking & 28.38 & 6.54 & 0.733 & 0.812 & 0.558 & 0.659 & 0.835 & 0.830 \\
Qwen3-VL-235B-Thinking & 30.37 & 6.55 & 0.742 & 0.815 & 0.571 & 0.674 & 0.852 & 0.825 \\
Qwen3-VL-8B-Inst & 24.31 & 6.05 & 0.732 & 0.801 & 0.557 & 0.655 & 0.888 & 0.825 \\
Llama-4-Scout-17B-Inst & 28.47 & 5.78 & 0.708 & 0.810 & 0.485 & 0.601 & 0.860 & 0.870 \\
Llama-4-Maverick-17B-Inst & 26.84 & 5.85 & 0.690 & 0.779 & 0.505 & 0.574 & 0.846 & 0.661 \\
MedGemma-27B-IT & 19.13 & 5.04 & 0.602 & 0.696 & 0.280 & 0.523 & \textbf{0.936} & 0.645 \\
MedGemma-1.5-4B-IT & 21.80 & 4.81 & 0.587 & 0.706 & 0.262 & 0.472 & 0.873 & 0.749 \\
MedGemma-4B-IT & 23.50 & 4.58 & 0.572 & 0.717 & 0.245 & 0.421 & 0.809 & 0.854 \\
\bottomrule
\end{tabular}}
\end{table*}

\section{Experiments and Discussion}
\label{sec:experiments_discussion}

\noindent \textbf{Experimental Setup:} We evaluate 16 vision-language models spanning closed-source systems (GPT-4o, o4-mini, GPT-5.2, Gemini-2.5/3-Pro/Flash), open-source foundation models (Qwen3-VL, InternVL3.5, Llama-4), and medically specialized VLMs (MedGemma). All models are evaluated zero-shot using multi-sequence, multi-view MRI inputs to assess out-of-the-box temporal reasoning capabilities. Closed-source models are accessed via APIs; open-source models are deployed locally. The public release of training and evaluation splits supports future supervised adaptation studies on this benchmark.

\noindent \textbf{Main Results Overview:} Table~\ref{tab:results} reports performance across final accuracy, RS, and TAC. Overall, models achieve moderate TAC (0.57--0.80) and chronology scores (0.64--1.00) but exhibit weaker Trend-F1 (0.19--0.63) and Sign Accuracy (0.42--0.74), indicating temporal alignment is more reliable than characterizing change direction. InternVL3.5-Inst leads on both TAC (0.80) and final accuracy (35.15\%), narrowly ahead of Gemini-3-Pro (35.10\%). Gemini-3-Flash shows lower accuracy (22.3\%) despite a perfect chronology score (1.00), suggesting potential instruction-following limitations or challenges with multi-sequence medical imaging. Specialized medical models (MedGemma variants) underperform general-purpose VLMs on TAC, indicating that domain specialization alone does not confer temporal reasoning ability. Collectively, results indicate that current VLMs handle temporal ordering reliably but consistently fail on change-type recognition, the clinically most critical capability.

\noindent \textbf{Multi-View Configuration Analysis:} To assess whether multi-view presentation compensates for the absence of volumetric processing in current VLMs, we conducted controlled ablations on the UCSF-GBM subset (1,192 samples) comparing multi-view (axial + coronal + sagittal) versus axial-only inputs. We employed an agentic Resident-Attending workflow: a Resident agent first extracts structured spatial findings (lesion location, signal changes, interval comparisons) from multi-timepoint MRI inputs, which an Attending agent then integrates for final diagnostic classification. Fixing the Resident prompt across view configurations reduces variation arising from prompt sensitivity. We evaluated representative models: GPT-4o, Gemini-2.5-Pro/Flash (closed-source) and Qwen3-VL-8B-Inst, MedGemma-4B-IT, InternVL3.5-Inst (open-source).

\noindent \textbf{Results:} Table~\ref{tab:agentic_results} reports accuracy on four clinical tasks. Multi-view inputs improve progression localization across models (average +6.2 pp) but produce mixed effects on temporal ordering: closed-source models benefit (GPT-4o: +7.2 pp), whereas smaller open-source VLMs show degradation (Qwen3-VL-8B: -8.0 pp, MedGemma-4B: -5.8 pp), suggesting information overload in compact architectures. Open-source models achieve highest global accuracy (InternVL3.5: 44.2\%, Qwen3-VL-8B: 43.6\%) versus closed-source models (GPT-4o: 36.7\%). Performance peaks on progression localization (97.3\%) and temporal ordering (72.0\%) but remains below 16\% on change segmentation and quantification, indicating that protocol-dependent segmentation and volumetric estimation remain fundamental bottlenecks independent of spatial localization success.

\begin{table}[t!]
\centering
\caption{Agentic workflow accuracy (\%) on UCSF-GBM subset across four clinical tasks. Multi-view inputs improve spatial localization but reveal persistent limitations in volumetric reasoning. Bold: best per column.}
\label{tab:agentic_results}
\scriptsize
\begin{tabular}{lccccc}
\toprule
\rowcolor{gray!15}
\textbf{Model} & \textbf{Global} & \textbf{Ch. Seg.} & \textbf{Ch. Quant.} & \textbf{Temp. Ord.} & \textbf{Prog. Loc.} \\
\midrule
\multicolumn{6}{c}{\textbf{Open-source models}} \\
\midrule
InternVL3.5-Inst & $\mathbf{44.2}$ & $1.1$           & $13.5$          & $\mathbf{72.0}$ & $\mathbf{97.3}$ \\
Qwen3-VL-8B-Inst & $43.6$          & $0.0$           & $12.7$          & $71.4$          & $96.9$          \\
MedGemma-4B-IT   & $42.1$          & $0.6$           & $11.7$          & $65.9$          & $96.2$          \\
\midrule
\multicolumn{6}{c}{\textbf{Closed-source models}} \\
\midrule
GPT-4o           & $36.7$          & $0.9$           & $12.7$          & $42.8$          & $95.1$          \\
Gemini-2.5-Flash & $27.8$          & $4.3$           & $9.2$           & $37.7$          & $63.3$          \\
Gemini-2.5-Pro   & $27.7$          & $\mathbf{15.7}$ & $\mathbf{15.5}$ & $44.2$          & $37.4$          \\
\bottomrule
\end{tabular}
\vspace{-1.0em}
\end{table}

\noindent \textbf{Implications:} Multi-view presentation improves spatial localization, demonstrating that orthogonal 2D planes provide complementary spatial context approaching clinical radiological workflows. However, view redundancy degrades temporal reasoning in smaller models, suggesting that adaptive view selection or attention mechanisms are needed to leverage multi-view inputs efficiently. The gap between strong localization (97\%) and limited volumetric quantification ($<$16\%) indicates that current VLMs lack the geometric reasoning required for protocol-adherent RANO measurements despite accurate spatial identification. Future architectures should integrate explicit 3D geometric priors, cross-timepoint attention, and protocol-conditioned training to bridge this gap.

\section{Conclusion}
We introduced the Time-Aware Multi-View MRI Benchmark for evaluating temporal reasoning in longitudinal MRI interpretation. Across 16 vision-language models, current systems achieve moderate temporal alignment but struggle with change characterization and volumetric reasoning, while multi-view inputs improve spatial localization at the cost of temporal coherence in compact architectures. Future work should prioritize native 3D architectures with explicit geometric priors and cross-timepoint attention to support longitudinal clinical decision-making.

\begin{credits}
\subsubsection{\discintname}
The authors have no competing interests to declare that are relevant to the content of this article.
\end{credits}

\bibliographystyle{splncs04}
\bibliography{bibliography.bib}

\end{document}